\documentclass{article} % For LaTeX2e
\usepackage{iclr2026_conference,times}

\usepackage{graphicx}
\usepackage[utf8]{inputenc} % allow utf-8 input
\usepackage[T1]{fontenc}    % use 8-bit T1 fonts
\usepackage{hyperref}       % hyperlinks
\usepackage{url}            % simple URL typesetting
\usepackage{booktabs}       % professional-quality tables
\usepackage{amsfonts}       % blackboard math symbols
\usepackage{nicefrac}       % compact symbols for 1/2, etc.
\usepackage{microtype}      % microtypography
\usepackage{xcolor}         % colors
\usepackage{svg}
\usepackage{amsmath}
\usepackage{multirow}
\usepackage{array}
\usepackage{geometry}
\usepackage{colortbl}
\usepackage{caption}
\usepackage{quoting}
\usepackage{subcaption}
\usepackage{listings}       % for code listings with line wrapping
\quotingsetup{leftmargin=1cm}
\usepackage{enumitem}
\usepackage[draft]{changes} % use 'draft' to show changes, 'final' to hide
\usepackage{float}

\usepackage{titlesec}
\titlespacing*{\subsection}{0pt}{1.1ex plus 0.3ex minus 0.1ex}{0.5ex plus 0.1ex}

\definechangesauthor[name={Yenchia Feng}, color=blue]{ps}

\iclrfinalcopy

\newcommand{\sect}[1]{\S\ref{#1}}

\makeatletter
\renewcommand\@makefnmark{\textsuperscript{\textdagger}}
\renewcommand\@fnsymbol[1]{\textdagger}
\makeatother

\title{Environment Maps: Structured Environmental Representations for Long-Horizon Agents}

\author{
Yenchia Feng \quad Chirag Sharma \quad Karime Maamari \\
Distyl AI \\
\texttt{\{yenchia, chirag.sharma, karime\}@distyl.ai} \\
}

\begin{document}
\maketitle

\begin{abstract}
Although large language models (LLMs) have advanced rapidly, robust automation of complex software workflows remains an open problem. In long-horizon settings, agents frequently suffer from cascading errors and environmental stochasticity; a single misstep in a dynamic interface can lead to task failure, resulting in hallucinations or trial-and-error. This paper introduces \emph{Environment Maps}: a persistent, agent-agnostic representation that mitigates these failures by consolidating heterogeneous evidence, such as screen recordings and execution traces, into a structured graph. The representation consists of four core components: (1) Contexts (abstracted locations), (2) Actions (parameterized affordances), (3) Workflows (observed trajectories), and (4) Tacit Knowledge (domain definitions and reusable procedures). We evaluate this framework on the WebArena benchmark across five domains. Agents equipped with environment maps achieve a 28.2\% success rate, nearly doubling the performance of baselines limited to session-bound context (14.2\%) and outperforming agents that have access to the raw trajectory data used to generate the environment maps (23.3\%). By providing a structured interface between the model and the environment, Environment Maps establish a persistent foundation for long-horizon planning that is human-interpretable, editable, and incrementally refinable.
\end{abstract}

\section{Introduction}
\label{sec:intro}

In spite of recent breakthroughs in large language models (LLMs), dependable automation of complex software workflows remains elusive. A core failure mode is long-horizon cascading error: agents may select reasonable local actions, yet fail to maintain a persistent understanding of (i) where they are in the interaction, (ii) which actions are valid, and (iii) what those actions will do over many steps \citep{chae2024webagents,gu2025isyourllm,wu2024oscopilot,yang2025learningonthejob}. This is exacerbated by UI drift and domain-specific semantics that differ from pretraining distributions \citep{li2024appagentv2,xie2025mirage1,xie2025guiexplorer,li2025adaptivemobile}.

Prior work tackles the problem of having autonomous agents solve long-horizon tasks in complex environments by introducing various structured representations. UI Transition Graphs capture navigation structure and reduce local trial-and-error \citep{wen2024autodroid,zhao2025llmexplorer,fang2025webevolver,wu2025rlvrworld}. Predictive models for web agents simulate short-term transitions from Document Object Model (DOM) observations and candidate actions, enabling limited look-ahead beyond reactive control \citep{chae2024webagents,gu2025isyourllm}. Separately, skill and workflow induction methods distill trajectories into reusable procedures or tools to shorten effective horizons \citep{wang2024agentworkflow,zheng2025skillweaver,wang2025inducing,prabhu2025walt,liu2025webcoach}. However, these approaches are typically developed in isolation from one another and often remain non-persistent or misaligned. Predictive models operate on raw observations rather than a shared abstract state space, skills lack explicit grounding in contexts and outcomes, and graphs can degrade under interface change. Consequently, agents still lack a unified, maintainable substrate that ties together \emph{state}, \emph{actions}, \emph{observed dynamics}, and \emph{task semantics} in a single artifact.

To bridge this gap, we introduce \emph{Environment Maps}, a persistent, agent-agnostic representation that consolidates heterogeneous experience (e.g., screen recordings, execution traces, structured workflows, documentation) into a queryable graph by segmenting workflows, clustering page contexts, and parameterizing actions e.g., ``Click John Smith'' $\rightarrow$ ``Click \texttt{\{customer\_name\}}''). Environment maps encode abstract contexts, parameterized actions, observed workflows, and tacit domain knowledge, separating durable environment knowledge from UI details and task-specific policies to support reuse across sessions and UI versions. In our experiments, agents equipped with environment maps achieve 28.2\% success on WebArena, nearly doubling the 14.2\% baseline rate of a \texttt{claude-sonnet-4-5} agent without map access, and outperform agents with access to the raw trajectories used to generate the maps (23.3\%), demonstrating that representation plays a significant role beyond raw data itself. Our contributions are as follows:
\begin{itemize}
    \item \textbf{A unified, agent-agnostic environment representation.}
    We introduce environment maps that integrate contexts, parameterized actions, workflows, and tacit knowledge into a single artifact, unifying ideas from UI graphs, skill libraries, and workflow memory.
    \item \textbf{Empirical demonstration of structured knowledge benefits.}
    Under controlled conditions with a fixed LLM-agent stack, we show that conditioning on environment maps yields higher success rates than baselines operating on raw observations or unstructured trajectory data.
    \item \textbf{An incrementally extensible design.}
    We propose a structured design that can be updated with new traces or expert edits, supporting ongoing maintenance as environments drift or workflows change.
\end{itemize}

\section{Environment Map Framework} \label{sec:envmap}

\subsection{Concept and Scope} \label{sec:envmap-concept}

We define an \emph{environment map} as a persistent, structured knowledge base that encodes what is known and what can be done within an environment. Formally, an environment map is a tuple $\mathcal{M} = (\mathcal{C}, \mathcal{A}, \mathcal{W}, \mathcal{K})$ where $\mathcal{C}$ is a set of \emph{contexts} (abstract locations), $\mathcal{A}$ is a set of \emph{actions} (parameterized affordances), $\mathcal{W}$ is a set of \emph{workflows} (observed trajectories), and $\mathcal{K}$ is \emph{tacit knowledge} (procedural logic and domain definitions). We additionally define $\mathcal{O}$ as raw observations (e.g., screen recordings, browser traces), $\mathcal{D}$ as interface representations (DOM or accessibility trees), and $\mathcal{A}'$ as unparameterized actions, used as intermediate representations in the creation pipeline.

An environment map (i) can adapt as new observations arrive, (ii) is interoperable for agents and humans (machine-readable yet human-editable), and (iii) supports multiple granularities and hierarchical groupings. Maps can be built from heterogeneous sources and, once constructed, provide a stable interface over a changing environment. Appendix~\ref{sec:appendix-structure} outlines the JSON schema for an environment map, along with visualizations.

\begin{figure}[t]
\centering
\includegraphics[width=0.9\columnwidth]{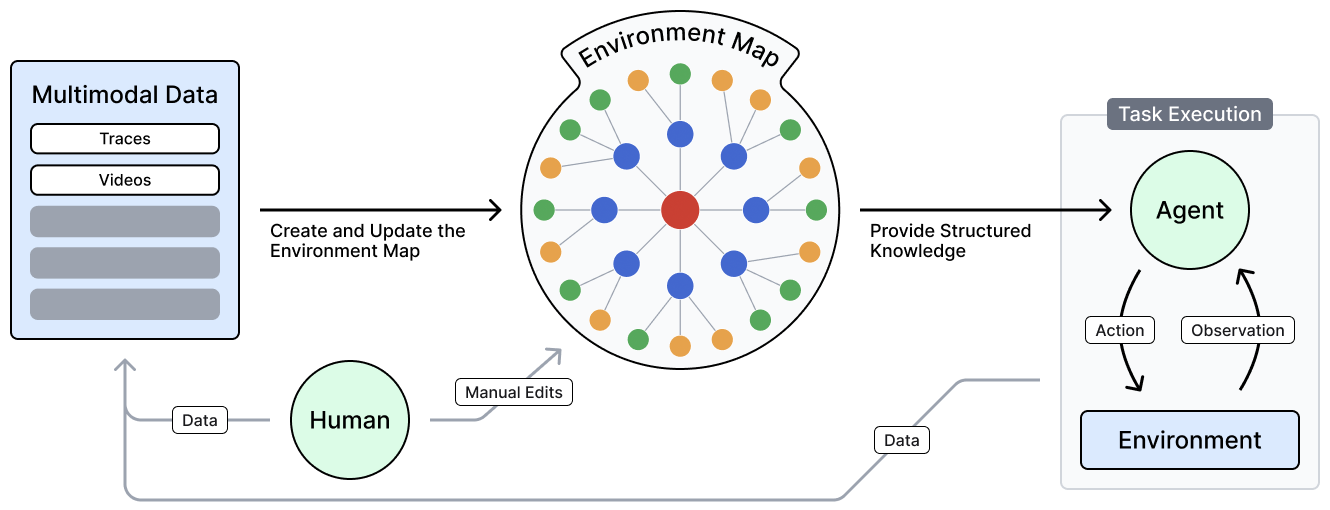}
\caption{\textbf{Environment map framework.} The system extracts structured spatial and semantic knowledge from multimodal data (e.g., traces and videos) to build an environment map. The map serves as a knowledge base for the agent during task execution, informing its action–observation loop. Solid black lines indicate pathways used in our experiments; gray lines indicate optional pathways not used, such as human-in-the-loop edits and map updates from agent traces, disabled to maintain a controlled experimental setting.}
\label{fig:framework}
\end{figure}

\subsection{Environment Map Creation Pipeline} \label{sec:envmap-pipeline}

The creation pipeline transforms raw observations $\mathcal{O}$ and interface representations $\mathcal{D}$ into maps $\mathcal{M}$ through five phases.

\paragraph{Phase 1: Unified Step Sequence}\mbox{}\\[0.5em]
A raw observation $o = (e_1, \ldots, e_T) \in \mathcal{O}$ is a sequence of events, where each $e_t$ records low-level interaction data (action type, timestamp, URL, element, utterance). Each observation is paired with an interface representation $d \in \mathcal{D}$ comprising DOM snapshots and screenshots that capture available UI affordances at each step. Phase 1 parses this into a workflow $w \in \mathcal{W}$ consisting of a step sequence $S = (s_1, \ldots, s_n)$, where each step $s_i = (n_i, d_i, r_i, u_i)$ comprises:
\begin{itemize}[nosep, label={}, leftmargin=1.5em]
    \item $n_i, d_i, r_i$: action name, description, and outcome (derived via LLM summarization of raw events and surrounding DOM context)
    \item $u_i$: the URL at which the step occurred
\end{itemize}

\paragraph{Phase 2: Action Extraction}\mbox{}\\[0.5em]
From workflows $\mathcal{W}$ and interface representations $\mathcal{D}$, Phase 2 extracts raw actions $\mathcal{A}' = \{a'_1, \ldots, a'_m\}$. Taken actions ($\tau = \text{taken}$) are extracted deterministically from step data, while potential actions ($\tau = \text{potential}$) are identified via LLM analysis of DOM elements not exercised in the trajectory. Each action $a'_i = (n_i, d_i, r_i, \tau_i, u_i)$ where:
\begin{itemize}[nosep, label={}, leftmargin=1.5em]
    \item $n_i, d_i, r_i$: action name, description, and expected outcome
    \item $\tau_i \in \{\text{taken}, \text{potential}\}$: whether the action was exercised in $\mathcal{W}$ or only available in the interface
    \item $u_i$: URL inherited from the source step (used for context assignment in Phase 4)
\end{itemize}

\paragraph{Phase 3: Action Generalization}\mbox{}\\[0.5em]
Phase 3 abstracts raw actions $\mathcal{A}'$ into parameterized actions $\mathcal{A}$ via LLM-based pattern detection. We define an equivalence relation $\equiv$ on $\mathcal{A}'$ where $a'_i \equiv a'_j$ iff they share the same interaction verb (e.g., Click, Type) and target element structure, differing only in specific values. Each equivalence class $[a'] \subseteq \mathcal{A}'$ yields a parameterized action $a \in \mathcal{A}$ where $a = (t, V)$ and:
\begin{itemize}[nosep, label={}, leftmargin=1.5em]
    \item $t$: action template with parameter placeholders (e.g., ``Click \texttt{\{link\_text\}}'')
    \item $V$: mapping from each parameter in $t$ to its observed values (e.g., $V(\texttt{link\_text}) = \{\text{``Settings''}, \text{``Profile''}\}$)
\end{itemize}

\paragraph{Phase 4: Context \& Knowledge Extraction}\mbox{}\\[0.5em]
Phase 4 produces contexts $\mathcal{C}$ and tacit knowledge $\mathcal{K}$. A \emph{context} represents a distinct page or view in the environment, identified by its URL pattern.

URL normalization is deterministic: regex-based rules replace dynamic segments with placeholders (e.g., \texttt{/users/123} $\rightarrow$ \texttt{/users/\{id\}}). Actions are grouped by their normalized URL $u_i$ to form contexts: each unique normalized URL yields a context $c \in \mathcal{C}$ where $c = (n_c, d_c, \lambda_c, A_c)$ and:
\begin{itemize}[nosep, label={}, leftmargin=1.5em]
    \item $n_c, d_c$: context name and description (LLM-generated from URL and page content)
    \item $\lambda_c$: the normalized URL pattern identifying when the agent is in context $c$
    \item $A_c \subseteq \mathcal{A}$: all parameterized actions whose source URL matches $\lambda_c$
\end{itemize}

Tacit knowledge $\mathcal{K} = (\mathcal{K}_D, \mathcal{K}_P)$ is LLM-extracted from trajectories and page content:
\begin{itemize}[nosep, label={}, leftmargin=1.5em]
    \item $\mathcal{K}_D$: domain definitions linking terminology to relevant actions and contexts (e.g., ``MR'' $\rightarrow$ ``Merge Request'', applicable to the merge request context)
    \item $\mathcal{K}_P$: procedures specifying common action sequences within a context (e.g., ``To filter issues: click Filter $\rightarrow$ select Label $\rightarrow$ choose value'')
\end{itemize}
See Listing~\ref{lst:schema} for the JSON encoding of contexts and Listing~\ref{lst:context} for action structure.

\paragraph{Phase 5: Integration \& Merging}\mbox{}\\[0.5em]
Phase 5 assembles the final environment map $\mathcal{M} = (\mathcal{C}, \mathcal{A}, \mathcal{W}, \mathcal{K})$ through deterministic merging:
\begin{itemize}[nosep, label={}, leftmargin=1.5em]
    \item Contexts with identical URL patterns $\lambda_c$ are merged, their action sets $A_c$ unioned
    \item Knowledge entries are deduplicated by ID; provenance links are preserved
    \item All entities carry stable IDs enabling cross-referencing and traceability to source observations $\mathcal{O}$
\end{itemize}

See Appendix \ref{sec:appendix-viz} for visualizations of the general environment map structure and a specific instance used in our experiments.

\section{Experiments}
\label{sec:experiments}

We evaluate on \textsc{WebArena}~\citep{zhou2023webarena}, using five interactive web environments from the benchmark: four core sites (an e-commerce store, a discussion forum, a GitLab-like code hosting platform, and an admin/CMS) plus a map site for geographic queries. WebArena provides 812 natural-language tasks across these environments, each requiring multi-step navigation and interaction rather than single-page actions. This setting directly stresses the capabilities that environment maps target: tracking entities and controls across pages and sites over long horizons. We compare agents equipped with environment maps against two baseline conditions and measure task completion success rates.

\subsection{Benchmark}
\label{sec:benchmark}

\paragraph{Environments.}
We evaluate across five WebArena environments: \textbf{E-Commerce} (Magento Storefront) for product search and checkout; \textbf{CMS} (Magento Admin) for order management and reports; \textbf{GitLab} (GitLab CE) for issue management and merge requests; \textbf{Reddit} (Postmill Forum) for posting and moderation; and \textbf{Map} (OpenStreetMap) for location search and routing. We evaluate on the complete task set for each environment (100--200 tasks per environment), spanning all intent templates and difficulty levels. We exclude Wikipedia tasks due to insufficient coverage in the human trajectory dataset.

\paragraph{Evaluation Protocol.}
We use the WebArena Verified evaluation framework~\citep{hattami2025webarena}, which provides deterministic, reproducible evaluation through HTTP Archive (HAR) trace inspection and type-aware task classification (RETRIEVE, MUTATE, NAVIGATE). In preliminary experiments comparing WebArena Verified against custom per-task evaluation functions (as in the original WebArena setup), we found that custom evaluators incorrectly passed some tasks due to formatting errors and agent hallucinations that were not caught without full HAR trace inspection.

\subsection{Experimental Conditions}
\label{sec:conditions}

We compare three experimental conditions that progressively add environment knowledge to the agent:

\paragraph{Baseline (No Map).}
The agent receives only the current page observation and task intent, with no prior knowledge of the environment structure. Observations are injected into the model prompt in the form of the accessibility tree for the webpage (see \sect{sec:agent}). This represents the typical setup for reactive web agents.

\paragraph{Trajectory Access.}
The agent has access to raw human trajectory data but \emph{not} the structured environment map. Human trajectories are stored as Playwright trace files in JSONL format containing low-level browser automation events including network requests, DOM snapshots, screencast frames, and action records with selectors and timestamps. The agent can query these files via CLI commands (e.g., \texttt{Read}, \texttt{Grep}) to find similar tasks and learn from human demonstrations. This condition isolates the value of raw trajectory data from the abstraction provided by environment maps.

\paragraph{Environment Map Access.}
The agent has access to the raw human trajectory data \emph{and} the full environment maps, one for each WebArena environment, constructed from the same raw human trajectory data using the pipeline from \sect{sec:envmap-pipeline}. The environment map includes page contexts (per-page action inventories), workflows (parameterized procedures from trajectory recordings), and tacit knowledge (domain definitions). The system prompt informs the agent of the map structure and provides usage patterns for consulting it on-demand via file access tools.

\subsection{Agent Configuration}
\label{sec:agent}

\paragraph{Base Agent.}
We use an agent built with the Claude Agent SDK, powered by \texttt{claude-sonnet-4-5}, with access to all default tools including file access (\texttt{Read}, \texttt{Grep}, \texttt{Glob}) and \texttt{Bash} utilities.

In initial experiments, we compared a Claude Agent SDK-based agent against GPT-5.2 with a custom agent loop. Both achieved similar task success rates on baseline evaluations, but the Claude agent offered faster execution and native support for file access tools, leading us to adopt it for all experiments.

We also found that disabling non-file tools significantly degraded baseline performance, as agents could no longer programmatically query web APIs or run analysis scripts. To benchmark realistic web agent performance near the state of the art, we therefore enable all default tools across conditions.

\paragraph{Observation and Action Space.}
The agent receives observations as accessibility trees extracted via Playwright \footnote{\textnormal{Playwright is a browser automation library: \href{https://playwright.dev/}{https://playwright.dev/}}}, following the standard WebArena observation format. To manage context window constraints, we apply filtering with three parameters: (i) consecutive sibling elements of the same role are collapsed after 10 items, preserving the first 10 with a summary indicator (e.g., \texttt{[...](row elements continue)}); (ii) individual text values exceeding 150 characters are truncated; and (iii) a hard 20,000 character cap is enforced. Across 122,523 observations in our baseline experiment, 8.0\% hit the character cap and 27.6\% had siblings collapsed, with environment-specific variation (e.g., 24.1\% truncation in CMS due to complex tables, vs. 0.2\% in E-Commerce). The median tree size was 4,281 characters, with 53.4\% of observations under 5,000 characters. This filtering preserves tree structure, element IDs, and the first 10 elements of each type while removing redundant list entries. The system prompt instructs the agent to output actions as structured JSON types (click, type, scroll, navigate, done) with target element identifiers from the accessibility tree.

\paragraph{Hyperparameters.}
We use consistent settings across all conditions: agents are limited to taking 30 steps in the environment per task and a 180-second timeout is enforced. Agents are permitted at most 10 conversation turns per step (includes reasoning and tool calls) and each turn response is limited to 4096 tokens. We use default SDK settings for model temperature and reasoning level.

\begin{table}[t]
\centering
\begin{tabular}{lcccc}
\toprule
Condition & Success (\%) & Mean steps (all) & Tool calls / task \\
\midrule
Baseline & $14.2$ \;[11.9,16.7] & $20.0$ \;[19.4,20.7] & $0.71$ (median $0$) \\
Trajectory access & $23.3$ \;[20.5,26.3] & $19.5$ \;[18.9,20.2] & $1.15$ (median $0$) \\
Environment Map & $28.2$ \;[25.2,31.4] & $18.5$ \;[17.8,19.1] & $12.06$ (median $9$) \\
\bottomrule
\end{tabular}
\caption{Overall performance across experimental conditions on WebArena ($n{=}812$ tasks). Brackets show Wilson 95\% confidence intervals. Steps are capped at 30 per task.}
\label{tab:overall_results}
\end{table}

\subsection{Environment Map Construction}
\label{sec:map_construction}

Environment maps are constructed following the pipeline described in \sect{sec:envmap-pipeline}.

\paragraph{Source Data.}
We use the human trajectory dataset released with WebArena, which contains 179 task recordings with a reported human success rate of 78.24\%. Each recording is a Playwright trace containing timestamped screenshots, user actions with Playwright selectors, page HTML snapshots, and network traffic.

\paragraph{Map Structure.}
For each WebArena environment, the constructed map includes: \textbf{12--77 page contexts} clustered by URL patterns with available actions and element descriptions; \textbf{19--45 workflows} derived from task recordings with step sequences and page references; and \textbf{tacit knowledge} comprising domain definitions ($\mathcal{K}_D$) and procedural hints ($\mathcal{K}_P$) as described in \sect{sec:envmap-pipeline}.

\paragraph{Coverage and Generalization.}
While the human trajectory dataset covers approximately 22\% of WebArena tasks, the environment maps are constructed to generalize beyond the source tasks: page contexts document UI structure applicable to any task navigating those pages, and workflows capture reusable navigation patterns and domain procedures.

\subsection{Implementation Details}
\label{sec:implementation}

WebArena Docker containers are hosted on cloud infrastructure and fully reset before each evaluation run to ensure reproducible starting conditions. All HAR traces are recorded for offline evaluation via WebArena Verified.

\section{Results}
\label{sec:results}

We report task success using the WebArena Verified evaluator (binary success score) on $812$ tasks spanning five single-site environments plus a multi-site subset. Unless otherwise stated, we report Wilson 95\% confidence intervals (CIs) for success rates and $t$-interval 95\% CIs for means (step counts, tool-call counts, and HAR-derived navigation metrics).

\subsection{Overall task success}
\label{sec:results-overall}

Table~\ref{tab:overall_results} summarizes end-to-end task success. The baseline agent succeeds on $14.2\%$ of tasks (CI $[11.9,16.7]$). Providing raw human trajectories improves success to $23.3\%$ (CI $[20.5,26.3]$), a relative gain of $64.3\%$ over baseline. Environment map access yields the best performance at $28.2\%$ (CI $[25.2,31.4]$), a $99.1\%$ relative gain over baseline and a $+4.9$ point gain over trajectories. These results indicate that transforming raw trajectories into an explicit environment map provides additional benefit beyond exposing the agent to those trajectories directly.

\begin{figure}[t]
\centering
\includegraphics[width=\columnwidth]{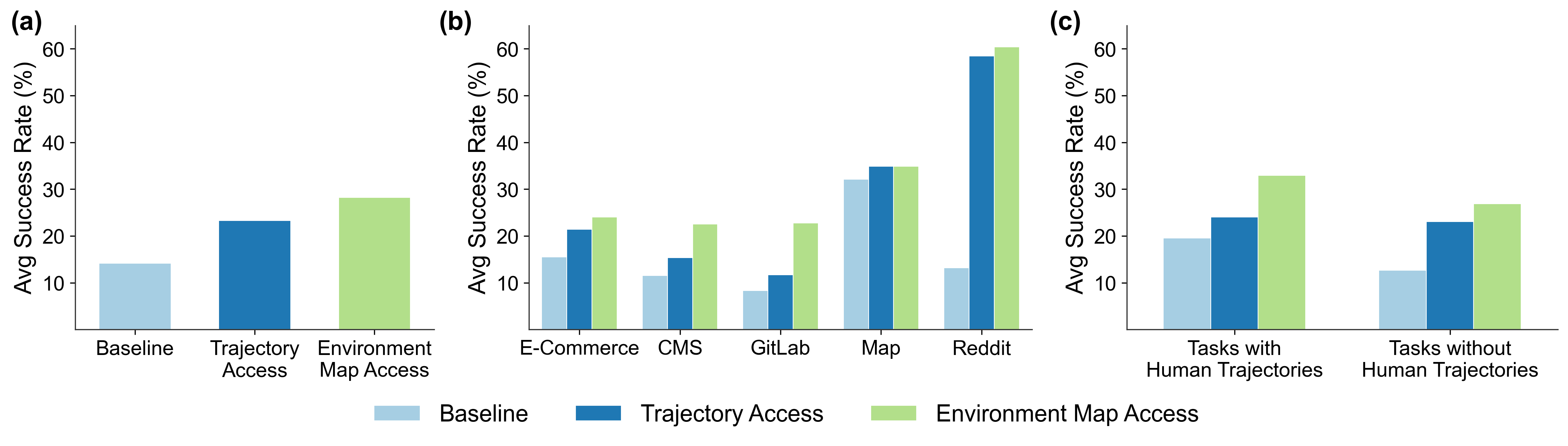}
\caption{\textbf{(a) Overall end-to-end task success on WebArena Verified.} Bars correspond to the three experimental conditions (Baseline, Trajectory access, Environment map access). $n{=}812$ tasks. \textbf{(b) Success rate by environment (single-site tasks).} The $x$-axis labels match the benchmark environments from \sect{sec:benchmark}: \emph{E-Commerce} denotes the Magento Storefront tasks ($n{=}187$), \emph{CMS} denotes the Magento Admin tasks ($n{=}182$), \emph{GitLab} denotes GitLab CE ($n{=}180$), \emph{Map} denotes OpenStreetMap ($n{=}109$), and \emph{Reddit} denotes the Postmill forum ($n{=}106$). \textbf{(c) Generalization vs.\ demonstration coverage.} Success rates on tasks that \emph{do} vs.\ \emph{do not} have a human demonstration trace (trace-covered: $n{=}179$; non-trace: $n{=}633$).}
\label{fig:results-overall}
\end{figure}

\subsection{Performance by environment}
\label{sec:results-by-env}

The impact of additional environment knowledge varies substantially across sites. Environment map yields the largest gains in UI-dense environments (\emph{GitLab}, \emph{CMS}).

Specifically, Figure~\ref{fig:results-overall}b shows that on \emph{Reddit}, success increases from $13.2\%$ in baseline to $58.5\%$ with trajectories and $60.4\%$ with the environment map, suggesting that for relatively low-branching interaction patterns, replayable demonstrations already capture much of the needed structure. In contrast, high-branching, UI-dense environments exhibit larger gaps between unstructured and structured knowledge. On \emph{GitLab}, success rises from $8.3\%$ to $11.7\%$ with trajectories, and further to $22.8\%$ with environment map. Similarly, on \emph{Content Management System (CMS)}, success increases from $11.5\%$ to $15.4\%$ and $22.5\%$. 

The \emph{Map} site shows a comparatively high baseline ($32.1\%$) and only modest gains with added context ($34.9\%$ for both trajectories and maps). Multi-site tasks remain largely unsolved (baseline $1/48$; trajectories and environment map $0/48$), indicating that per-site representations do not help with solving cross-site workflows in this benchmark setting.

\subsection{Generalization beyond demonstrated tasks}
\label{sec:results-generalization}

To measure generalization, we partition tasks by whether the task has an associated human trajectory in the released trace set ($n{=}179$ with trajectory traces; $n{=}633$ without). Figure~\ref{fig:results-overall}c shows that on trace-covered tasks, success improves from $19.6\%$ to $24.0\%$ with trajectories and $33.0\%$ with environment maps. Importantly, gains persist on tasks \emph{without} direct demonstrations: baseline achieves $12.6\%$, trajectories $23.1\%$, and environment map $26.9\%$. Thus, while raw demonstrations help most when they closely match solutions to the target task, they also retain an advantage when the agent is generalizing to new tasks in the environment. The structured environment map generated from these demonstrations further improves generalization performance, suggesting that it enables more targeted retrieval of environment knowledge.

\begin{figure}[t]
\centering
\includegraphics[width=0.8\columnwidth]{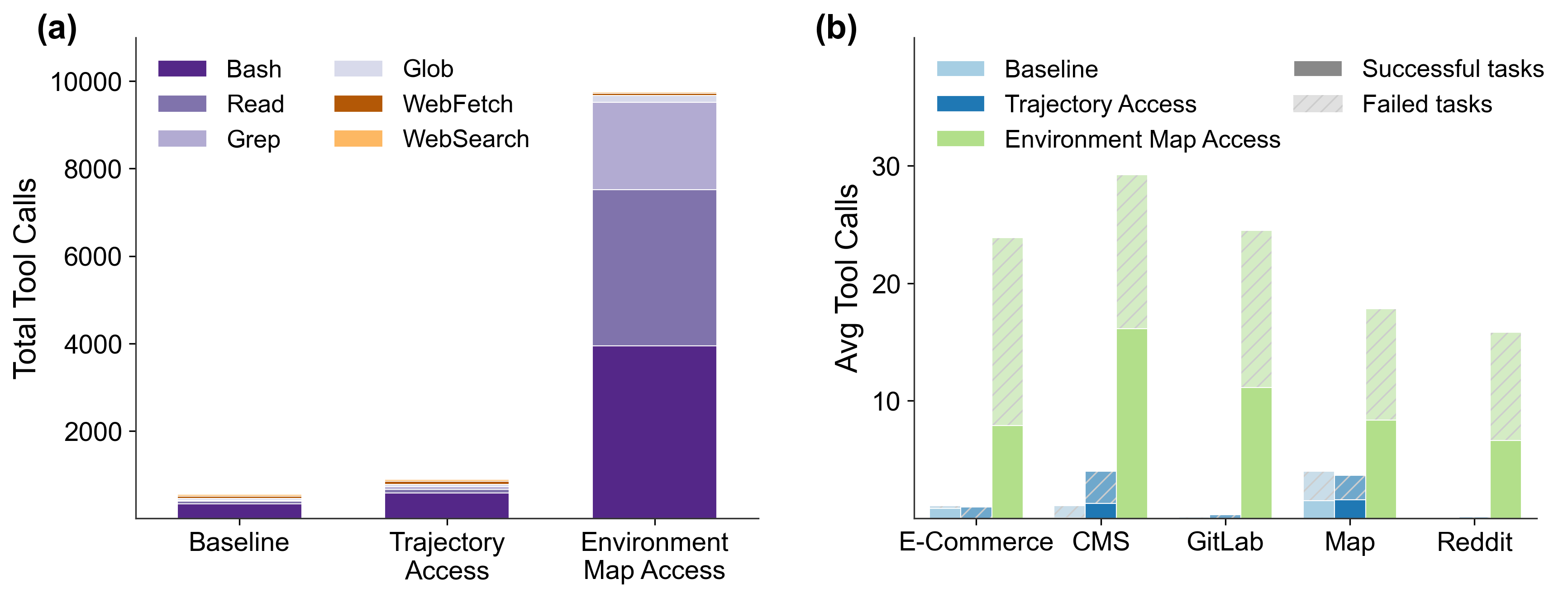}
\caption{\textbf{(a) Aggregate tool usage by experimental condition.} Stacked bars show total tool calls by type across all $n{=}812$ tasks. \textbf{(b) Mean tool calls per task by environment and outcome.} Grouped bars show average tool calls for each condition, split by task outcome and environment.}
\label{fig:results-tools}
\end{figure}

In the environment map condition, a majority of the tool calls made by the agent are file-related: \texttt{Read} (3566) + \texttt{Grep} (2001) + \texttt{Glob} (162) account for 5729 calls. All other tools combined (including \texttt{Bash} and web tools such as \texttt{WebFetch}/\texttt{WebSearch}) account for 4066 calls, indicating that the agent spends a majority of its tool interactions on local file inspection of map artifacts. Notably, Figure~\ref{fig:results-tools}b shows the two environments with highest tool usage for successful tasks, CMS and GitLab, also show the largest performance gains from trajectory to environment map access (Figure~\ref{fig:results-overall}b), suggesting that increased map consultation in complex environments translates to improved success rather than wasted exploration.

\subsection{HTTP Archive (HAR)-based navigation analysis}
\label{sec:results-har}

We analyze HAR traces to characterize exploration and backtracking. On the subset with available HAR logs (baseline $n{=}611$; trajectory and environment map $n{=}811$), the baseline exhibits a high backtracking rate (mean $0.445$). Both trajectory and environment map access reduce backtracking (means $0.376$ and $0.381$ respectively), and total network requests are lower with the environment map (mean $707$ vs.\ $892$ in baseline).

\section{Discussion}
\label{sec:discussion}

\subsection{Summary of Findings}
Our experiments show that environment maps provide an effective, agent-agnostic substrate for automation. Replacing either no prior knowledge or raw trajectories with environment maps (while keeping the underlying agent stack fixed) improves success rates and leads to more targeted use of external knowledge, with the largest gains in complex, high-branching environments. These improvements require no policy retraining, and the resulting maps remain interpretable and human-editable.

\subsection{Portability and Maintenance}
Environment maps are constructed independently of any specific model or prompting strategy and serve solely as an external information source. This separation yields two practical benefits: (i) updated models can reuse existing maps without retraining, and (ii) maps can be incrementally maintained as environments evolve using new trajectories or expert edits, without changing the policy. In our setting, constructing maps for the five WebArena environments used between 19 (Reddit) and 45 (E-Commerce) trajectories at a cost of approximately \$1 to \$4 and around 13 to 31 minutes per environment. Applying these maps to 812 evaluation tasks yielded 114 additional successes. Even relatively lightweight supervised fine-tuning of a mid-sized LLM typically consumes on the order of tens to hundreds of GPU-hours. By contrast, our maps incur a small, one-off construction cost and then remain reusable across models and prompting strategies without any additional training.

\subsection{Representation Matters: Maps vs. Raw Trajectories and Contextualization}
\label{subsec:representation_matters}
Providing the agent with raw human trajectory traces increased success from 14.2\% to 23.3\%. However, restructuring that same underlying data into an environment map further improved success to 28.2\%. This indicates that the utility of prior experience is significantly enhanced by its organization. While raw traces are difficult to query and mask cross-session patterns, environment maps distill them into a compact, actionable hierarchy. The largest gains occurred in high-branching environments like \emph{GitLab} and \emph{CMS}, where explicit navigation and action schemas make distributed knowledge reusable. To isolate the map's impact, we held the retrieval interface fixed. Agents used default file tools (e.g., \texttt{Read}, \texttt{Grep}) rather than custom retrieval pipelines, ensuring that performance gains are driven by the representation itself.

\subsection{Interaction Patterns and Interpretability}
Differences in representation are reflected in tool use. With environment maps, agents make an average of 12.1 tool calls per task (primarily file reads and searches), and 92\% of tasks involve file access. Under the raw trajectory setting, agents make only 0.2 file calls per task, and only 3.2\% of tasks involve file access. This disparity reflects the usability of each format: environment maps are structured JSON with clear indices, while raw trajectories are Playwright trace recordings with hundreds of files per task in a format not designed for LLM consumption. Because maps are structured text with simple indices, they are also human-interpretable and editable. Operators can inspect the structure, diagnose misclassifications, and apply targeted edits. This transparency contrasts with latent representations that are difficult to audit or modify directly.

\subsection{Limitations and Future Work}
The utility of environment maps varies with site complexity. On simpler environments like \emph{Reddit}, where navigation is shallow, raw trajectories provide comparable benefits; on high-branching systems like \emph{GitLab} and \emph{CMS}, environment maps yield substantially larger gains. Future work should investigate the threshold at which structured maps become strictly superior to raw traces. While our schema includes action preconditions and effects, these are currently populated descriptively---transitioning to grounded state observations would enable real-time validation. Additionally, our current maps are restricted to single environments; supporting cross-site workflows and shared entity abstractions remains unaddressed. Finally, as web environments evolve, maps may become stale; developing methods to automatically detect and repair structural errors with minimal human intervention is a critical area for future research.

\subsection{Conclusion}
Environment maps provide a compact, human-readable, and agent-agnostic intermediate representation. Compared to naive baselines and raw demonstrations, they preserve low-level multimodal signals while imposing a structure that facilitates targeted lookups by agents and enables human inspection. In complex, high-branching web interfaces, these properties enable substantial improvements in success and efficiency.

\bibliographystyle{iclr2026_conference}
\bibliography{iclr2026_conference}
\clearpage

\appendix
\section{Environment Map Structure}
\label{sec:appendix-structure}

This appendix provides a concrete illustration of the environment map representation. We first present the JSON schema (generalized template and truncated GitLab instance), then visualize the hierarchical structure.

\subsection{JSON Schema}
\label{sec:appendix-json}

\paragraph{Generalized Template.}
An environment map $\mathcal{M}$ is serialized as a JSON object with the following top-level structure:

\begin{lstlisting}[language={}, basicstyle=\small\ttfamily, frame=single, caption={Top-level environment map JSON schema. The \texttt{page\_contexts} array contains context summaries with references to detailed context files; \texttt{workflows} captures task-specific action sequences; \texttt{tacit\_knowledge} stores domain definitions and procedural hints; \texttt{statistics} and \texttt{metadata} provide provenance information.}, label={lst:schema}]
{
  "id": "<environment-id>",
  "name": "<Environment Name>",
  "description": "<Human-readable description>",
  "base_url": "<base URL pattern>",
  "page_contexts": [
    {
      "context_id": "context.<name>",
      "name": "<context_name>",
      "description": "<Context description>",
      "pattern": "<URL pattern, e.g., /search>",
      "context_mesh_path": "<path to detailed context file>",
      "action_count": <N>,
      "contributing_recordings": ["task_1", "task_2", ...]
    },
    ...
  ],
  "workflows": [
    {
      "workflow_id": "workflow.<name>",
      "workflow_mesh_path": "<path to detailed workflow file>"
    },
    ...
  ],
  "tacit_knowledge": {
    "definitions": [...],
    "procedures": [...]
  },
  "statistics": { ... },
  "metadata": { ... }
}
\end{lstlisting}

\clearpage
Each \texttt{page\_context} references a detailed context file containing parameterized actions:

\begin{lstlisting}[language={}, basicstyle=\small\ttfamily, frame=single, caption={Detailed context file structure showing parameterized actions. Each action includes a generalized template (e.g., ``Click link''), parameter metadata for instantiation, and \texttt{instances} tracking which specific values were observed in trajectories. The \texttt{provenance} field links each instance back to its source task and step number.}, label={lst:context}]
{
  "id": "context.<name>",
  "pattern": "<URL pattern>",
  "available_actions": [
    {
      "action": "<Action template, e.g., Click link>",
      "action_id": "action.<id>",
      "type": "generalized",
      "is_parameterized": true,
      "parameter_name": "<param, e.g., link_text>",
      "possible_values": ["value1", "value2", ...],
      "instances": [
        {
          "action_id": "instance.<task>_<step>_<idx>",
          "is_taken": true,
          "action_description": "<Specific action taken>",
          "provenance": {
            "source": "<trajectory source>",
            "task_id": "<task_id>",
            "step_number": <N>
          }
        },
        ...
      ]
    },
    ...
  ]
}
\end{lstlisting}

\clearpage
\paragraph{GitLab Instance (Truncated).}
The following shows the actual GitLab environment map structure:

\begin{lstlisting}[language={}, basicstyle=\small\ttfamily, frame=single, caption={Truncated GitLab environment map instance derived from 41 WebArena task trajectories. This map identifies 96 distinct page contexts (e.g., main dashboard, search results, project pages) containing 491 total actions. Each context tracks which tasks contributed observations via \texttt{contributing\_recordings}, enabling provenance tracing. Only two representative contexts are shown; the full map spans diverse GitLab functionality including issues, merge requests, user profiles, and repository browsing.}, label={lst:gitlab}]
{
  "id": "env-map-gitlab",
  "name": "Gitlab Environment Map",
  "description": "Merged environment map from 41 WebArena gitlab tasks",
  "base_url": "__GITLAB__",
  "page_contexts": [
    {
      "context_id": "context.gitlab_main",
      "name": "gitlab_main",
      "description": "The main page on gitlab.",
      "pattern": "/",
      "action_count": 512,
      "contributing_recordings": ["task_791", "task_259", ...]
    },
    {
      "context_id": "context.gitlab_search",
      "name": "gitlab_search",
      "description": "The search page on gitlab.",
      "pattern": "/search",
      "action_count": 56,
      "contributing_recordings": ["task_668", "task_686", ...]
    },
    ... // 94 more contexts
  ],
  "workflows": [
    {"workflow_id": "workflow.gitlab_task_103", ...},
    {"workflow_id": "workflow.gitlab_task_135", ...},
    ... // 39 more workflows
  ],
  "statistics": {
    "num_steps": 491,
    "pages_identified": 96,
    "actions_extracted": 491,
    "recordings_processed": 41
  }
}
\end{lstlisting}

\clearpage
\subsection{Visualization}
\label{sec:appendix-viz}

The following figures illustrate the hierarchical structure of environment maps using a concentric layout: the environment node (red) at center, context nodes (blue) in an inner ring, and action nodes in the outer ring. Orange nodes represent \emph{taken actions} observed in recorded trajectories; green nodes represent \emph{potential actions} inferred through parameterization. Nodes with a purple glow contain associated tacit knowledge. Blue edges indicate \emph{sequential} relationships (actions commonly performed in succession); purple edges indicate \emph{alternative} relationships (mutually exclusive actions within the same context).

\begin{figure}[h]
\centering
\includegraphics[width=0.99\textwidth]{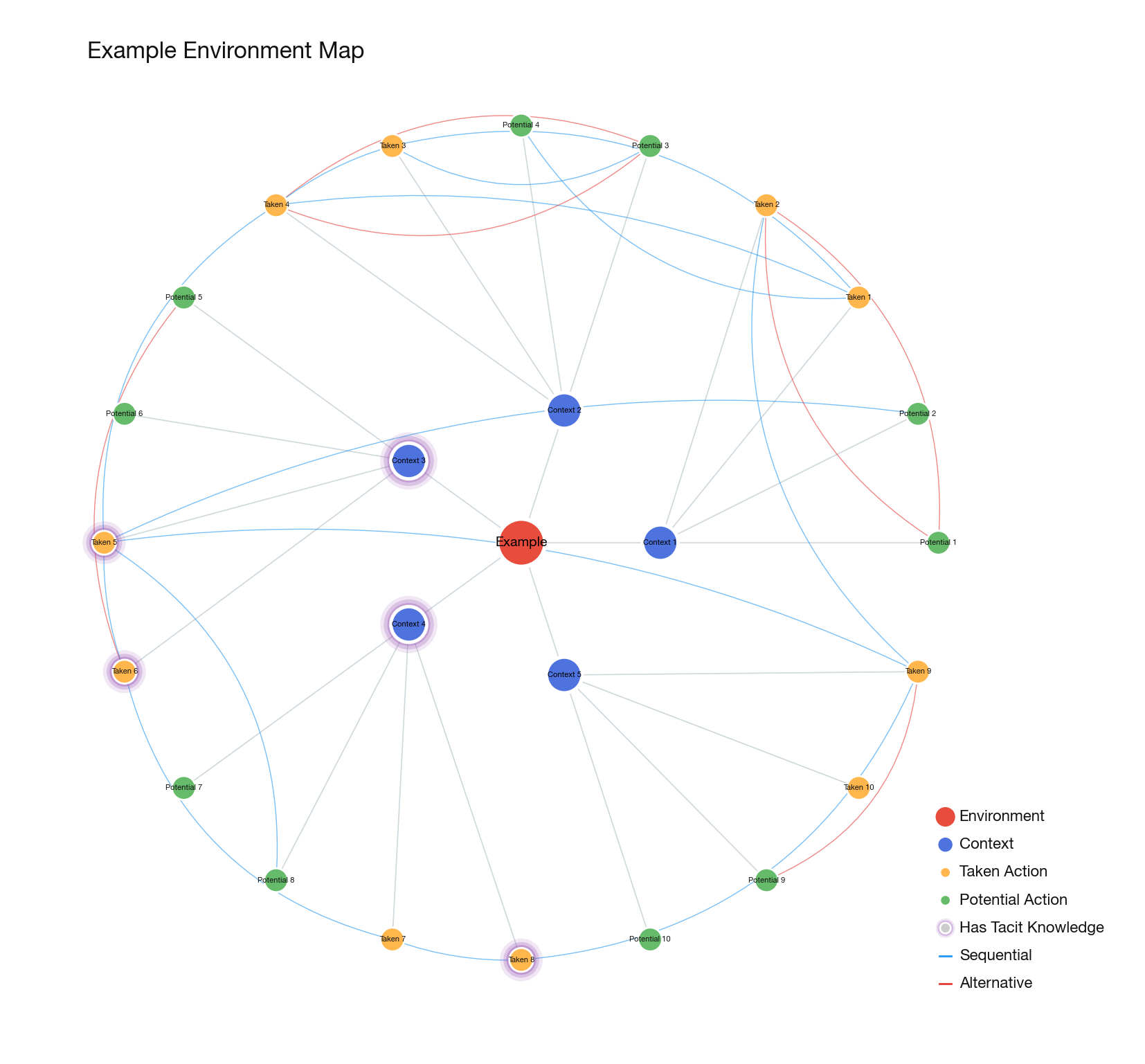}
\caption{\textbf{Generalized environment map structure.} A synthetic map example demonstrating the various node types and edge relationships.}
\label{fig:viz-generalized}
\end{figure}

\clearpage
\begin{figure}[h]
\centering
\includegraphics[width=0.99\textwidth]{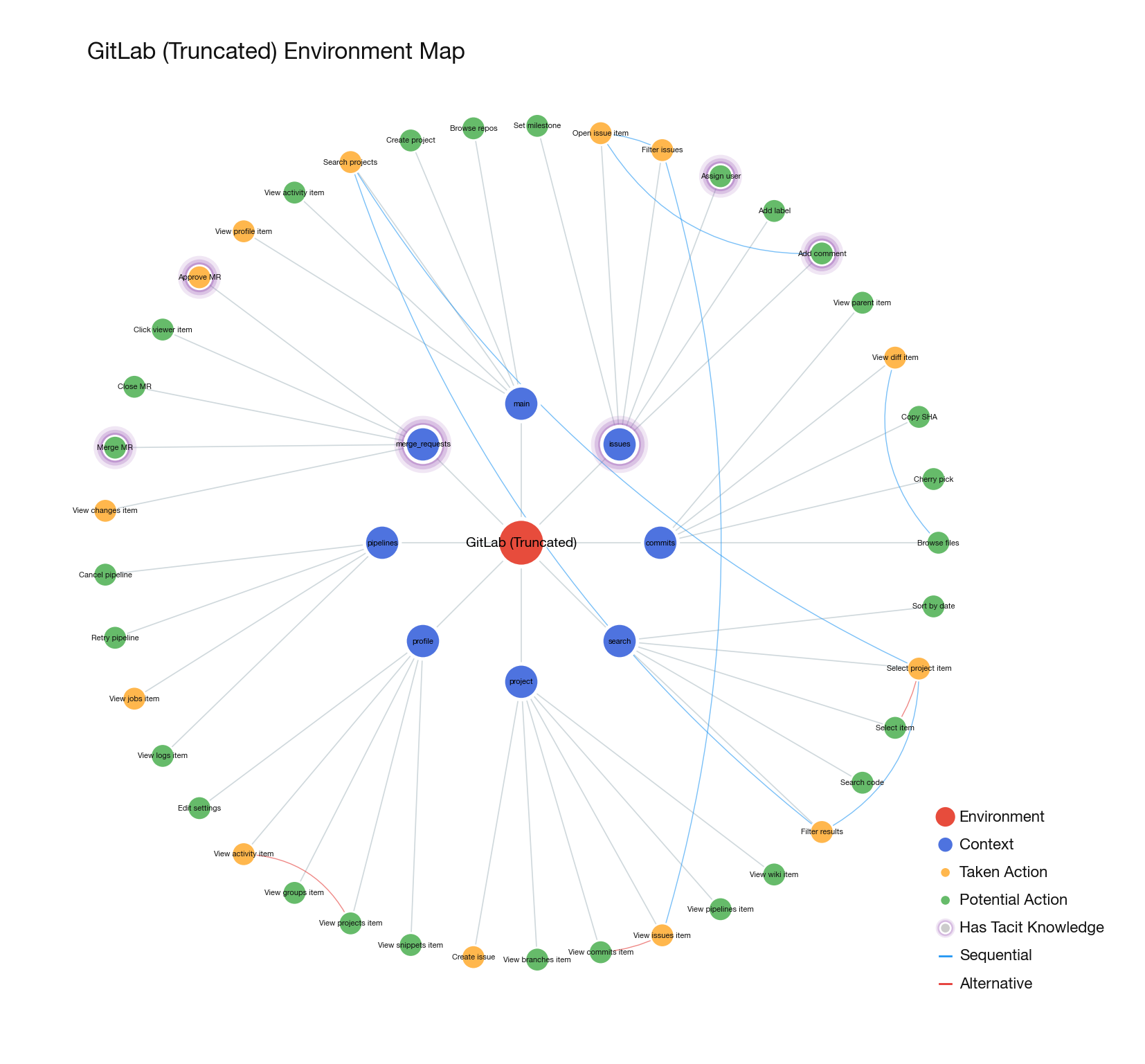}
\caption{\textbf{GitLab environment map (curated subset).} A curated 8-context, 42-action subset of the full 96-context GitLab environment map derived from 41 WebArena task trajectories. This visualization shows representative contexts including the main dashboard, search page, project views, and user settings. The density of connections illustrates the navigational structure captured from real agent interactions, while the mix of taken (orange) and potential (green) actions demonstrates how parameterization expands the action space beyond observed instances.}
\label{fig:viz-gitlab}
\end{figure}

\end{document}